\documentclass[runningheads]{llncs}
\usepackage{nips13submit_e,times}
\usepackage{graphicx}

\usepackage{color}
\usepackage{subfigure}
\usepackage{tabularx}
\usepackage{multirow}
\usepackage{times}
\usepackage{epsfig}
\usepackage{soul}
\newcommand{\tabincell}[2]{\begin{tabular}{@{}#1@{}}#2\end{tabular}}

\usepackage{graphicx}
\usepackage{amsmath,amssymb} % define this before the line numbering.
\usepackage{color}
\nipsfinalcopy % Uncomment for camera-ready version
\begin{document}
\title{Orthogonal Deep Features Decomposition for Age-Invariant Face Recognition} 
% Replace with your title

\titlerunning{Orthogonal Deep Features Decomposition for Age-Invariant Face Recognition}
% Replace with a meaningful short version of your title
%
%\author{First Author\inst{1}\orcidID{0000-1111-2222-3333} \and
%Second Author\inst{2,3}\orcidID{1111-2222-3333-4444} \and
%Third Author\inst{3}\orcidID{2222--3333-4444-5555}}
%\author{Yitong Wang, Dihong Gong, Zheng Zhou, Xing Ji, Hao Wang, \\Zhifeng Li\thanks{Corresponding authors}, Wei Liu\samethanks {} and Tong Zhang}
\author{
  Yitong Wang \quad Dihong Gong \quad Zheng Zhou \quad Xing Ji \quad Hao Wang \quad Zhifeng Li\thanks{Corresponding author} \quad Wei Liu\thanks{Corresponding author} \quad Tong Zhang\\
  Tencent AI Lab, China\\
  \texttt{\{yitongwang,dihonggong,encorezhou,denisji,hawelwang,michaelzfli\}@tencent.com} \\
    \texttt{wl2223@columbia.edu \quad tongzhang@tongzhang-ml.org} \\
  %% \AND
  %% Coauthor \\
  %% Affiliation \\
  %% Address \\
  %% \texttt{email} \\
  %% \And
  %% Coauthor \\
  %% Affiliation \\
  %% Address \\
  %% \texttt{email} \\
  %% \And
  %% Coauthor \\
  %% Affiliation \\
  %% Address \\
  %% \texttt{email} \\
}
%
%Please write out author names in full in the paper, i.e. full given and family names. 
%If any authors have names that can be parsed into FirstName LastName in multiple ways, please include the correct parsing, in a comment to the volume editors:
%\index{Lastnames, Firstnames}
%(Do not uncomment it, because you may introduce extra index items if you do that, we will use scripts for introducing index entries...)
\authorrunning{Y. Wang et al.}
% Replace with shorter version of the author list. If there are more authors than fits a line, please use A. Author et al.
%

\institute{Tencent AI Lab, China \\
\email\{yitongwang,encorezhou,denisji,hawelwang,michaelzfli\}@tencent.com \\
gongdihong@gmail.com
wl2223@columbia.edu
tongzhang@tongzhang-ml.org
}
\maketitle              % typeset the header of the contribution
\begin{abstract}
As facial appearance is subject to significant intra-class variations caused by the aging process over time,
age-invariant face recognition (AIFR) remains a major challenge in face recognition community. To reduce the intra-class discrepancy caused by the aging, in this paper we propose a novel approach (namely, Orthogonal Embedding CNNs, or OE-CNNs) to learn the age-invariant deep face features. Specifically, we decompose deep face features into two orthogonal components to represent age-related and identity-related features. As a result, identity-related features that are robust to aging are then used for AIFR. Besides, for complementing the existing cross-age datasets and advancing the research in this field, we construct a brand-new large-scale Cross-Age Face dataset (CAF).
Extensive experiments conducted on the three public domain face aging datasets (MORPH Album 2, CACD-VS and FG-NET) have shown the effectiveness of the proposed approach and the value of the constructed CAF dataset on AIFR. Benchmarking our algorithm on one of the most popular general face recognition (GFR) dataset LFW additionally demonstrates the comparable generalization performance on GFR.

\end{abstract}

\section{Introduction}

\begin{figure*}[t]
  \centering
  % Requires \usepackage{graphicx}
  %An example case which 类间距离 小于 类内距离
  \subfigure[]
  {
    \label{fig:1:a}
    \includegraphics[width=0.5\linewidth, keepaspectratio]{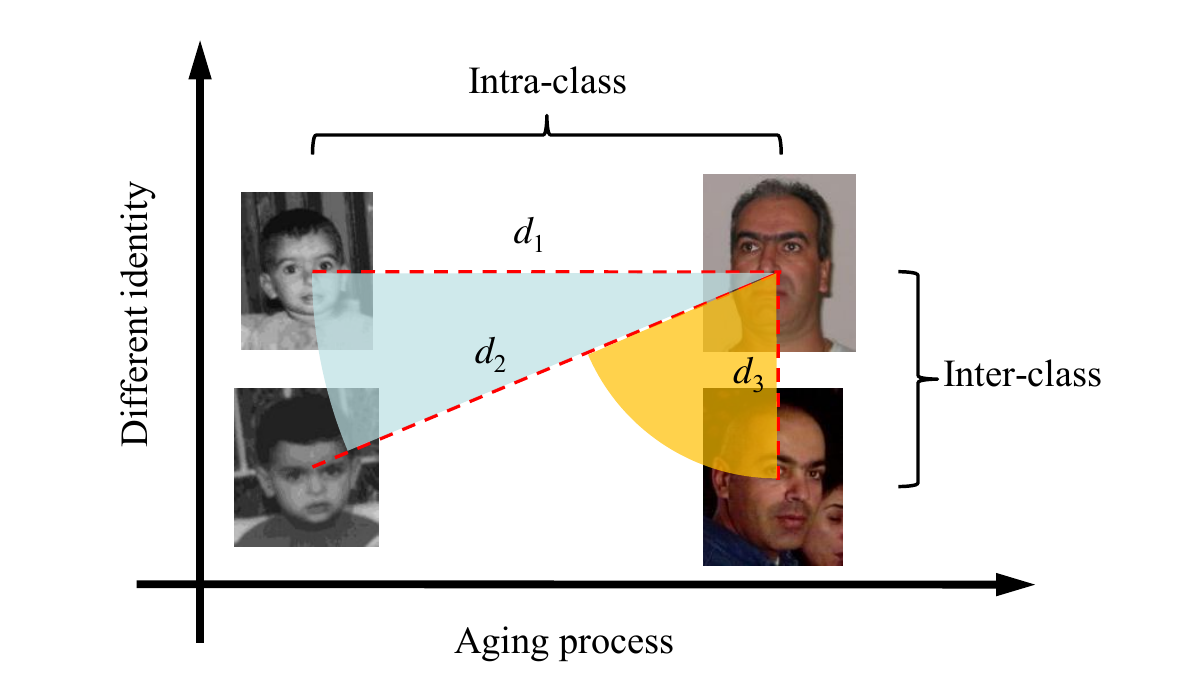}
  }
  \subfigure[]
  {
    \label{fig:1:b}
    \includegraphics[width=0.45\linewidth, keepaspectratio]{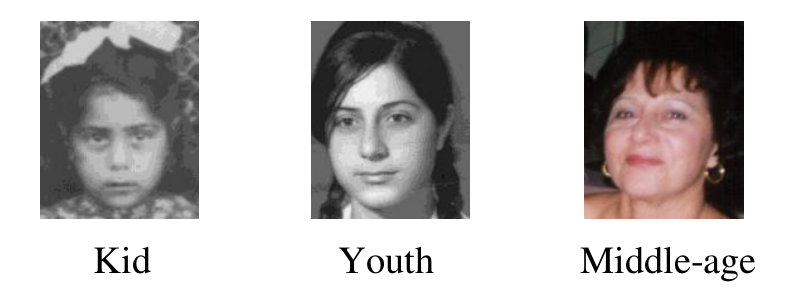}
  }
  %The major challenge of AIFR. The large intra-class variations 使得识别人脸的难度大大提高。
  \caption{The major challenge of AIFR: the intrinsic large intra-class variations in the aging process. (a) An example where intra-class distance is larger than inter-class distance. (b) The cross-age images for one subject in the FG-NET \cite{fgnet}.}
  \label{fig:1}
\end{figure*}

As one of the most important topics in computer vision and pattern recognition, face recognition has attracted much attention from both academic and industry for decades \cite{tian2007face,turk1991face,Belhumeur:1997,ahonen2006face,wang2004unified,Li05nonparametricsubspace,Li2009NonparametricDA,spatio-temporal,Xiong:2013,liu2004null,liu2004nullkernel}. With the evolution of deep learning, the performance of general face recognition (GFR) has been significantly improved in recent years, even higher than humans' abilities \cite{deepid2,deepid2plus,deepid3,facenet,centerloss,sphereface,cosface}. As a major challenge in face recognition, age-invariant face recognition (AIFR) is extremely valuable on various application scenarios, such as looking for lost children after decades, matching face images in different ages, etc. In contrast to GFR, AIFR involves more diversity with the significant intra-class variations caused by the aging process and thus is more challenging. It is very often that the inter-class variation is much smaller than the intra-class variation in the presence of age variation, as illustrated in Figure \ref{fig:1:a}. Figure \ref{fig:1:b} also exhibits the difficulty of AIFR where the same identity greatly varies in appearance with the aging process.

Recent AIFR researches primarily concentrate on two technical schemes: generative scheme and discriminative scheme. The generative scheme models the AIFR by synthesizing faces to one or more fixed age category then performs recognition with the artificial face representations \cite{g1,g2,g3}. Benefited from the advancement of the deep generative model, the generative scheme becomes more promising on AIFR as well  \cite{caae,cgan,TNVP}.
However, the generative scheme still remains several significant shortcomings.
Firstly, generative scheme usually separates the recognition process into two steps. Hence it is not easy for the generative models to optimize recognition performance in an end-to-end manner.
Secondly, generation models are often unstable so the synthesizing face images will introduce additional noises, which may result in negative effects on the recognition process.
Moreover, constructing an accurate, parametric generation model is fairly difficult since the aging process of humans' face is easily impacted by many latent factors such as social environments, diet, etc.

The discriminative scheme aims at constructing the sophisticated discriminative model to solve the problem of AIFR. Related works on discriminative model include \cite{mefa,lps_hfa,gsm,chen2013blessing,d1,d2,cacd,cacd2,hfa}. By combining the deep learning algorithm, the discriminative scheme has achieved substantial improvement on AIFR.
For example, Wen et al. \cite{LFCNN} extended the HFA method \cite{hfa} to a deep CNN model called latent factor guided convolutional neural networks (LF-CNNs), which achieved the state-of-the-art recognition accuracy in this field.
Zheng et al. \cite{AECNN} also used the linear combination of jointly-learned deep features to represent identity and age information, which is similar to the HFA based deep CNN model.

%Despite the combination of CNN models and discriminative approaches has resulted in a tremendous improvement, how to introduce new deep learning methods (such as hypersphere based methods \cite{l2softmax,lsoftmax,sphereface,normface,cocoloss_v2}) to AIFR, is still an open problem.

In this paper, we aim at designing a new deep learning approach to effectively learn age-invariant components from features mixed with age-related information. The key idea of our approach is to decompose face features into age-related and identity-related components, where the identity-related component is age-invariant and suitable for AIFR. More specifically, inspired by a recent state-of-the-art deep learning GFR system with A-Softmax loss \cite{sphereface} where features of different identities are discriminated by different angles, we decompose face features in the spherical coordinate system which consists of radial coordinate $r$ and angular coordinates $\phi_1,\dots,\phi_{n}$. Then the identity-related components are represented with angular coordinates, and the age-related information is encoded with radial coordinate. Features separated by the two mutually orthogonal coordinate systems are then trained jointly with different supervision signals. Identity-related features are trained as a multi-class classification task supervised by identity labels with the A-Softmax loss, and age-related features are trained as a regression task supervised by age labels. As such, we extract age-invariant features from angular coordinates by separating age-related components with radial coordinates. Since face features are decomposed into mutually orthogonal coordinate systems, we name our approach as orthogonal embedding CNNs (OE-CNNs). 
A related work Decoupled Network also discussed how to decouple the CNN with orthogonal geometry in details. Nevertheless, this work merely studies the generalization of networks rather than specifically modeling the age into decomposed features in the AIFR application scenario.
%TODO: A similar work Decoupled Networks \cite{} 同样研究了如何将CNN基于几何正交解耦的问题，不过该工作是从网络泛化性角度研究的，并未结合age-invariant face recognition的应用场景将年龄信息引入到norm的建模中。
We verify the effectiveness of OE-CNNs with extensive experiments on three face aging datasets (MORPH Album2 \cite{morph}, CACD-VS \cite{cacd} and FG-NET \cite{fgnet}) and one GFR dataset (LFW \cite{lfw}), and achieve the state-of-the-art performances.

The major contributions of this paper are summarized as follows:

1. We propose a new approach called OE-CNNs to tackle the problem on how to jointly model the age-related features and identity-related features in a deep CNN model.
Based on the proposed model, age-invariant deep features can be effectively obtained for improved AIFR performance.

2. We introduce a new large-scale Cross-Age Face dataset, named CAF, to help advance the research in this field. This dataset contains more than 313,986 images from 4,668 identities. The face data in CAF has been manually cleaned in order to be noise-free.

%The dataset characterizes to be manually cleaned, large-scale and nearly noise-free, which contains more than 313,986 images with 4,668 identities.

3. We demonstrate the effectiveness of our proposed approach with several extensive experiments over three face aging datasets (MORPH Album2 \cite{morph}, CACD-VS \cite{cacd} and FG-NET \cite{fgnet}) and one GFR dataset (LFW \cite{lfw}). The experimental results have shown the superior performance of the proposed approach over the state-of-the-art either on AIFR or GFR.

\section{Proposed Approach}

\subsection{Orthogonal Deep Features Decomposition}

Two certain difficulties involved in AIFR include the considerable variations of the identical individual in different age categories (intra-class variations) caused by aging process (such as shape changes, texture changes, etc.), and the inevitable mixture of unrelated components in the deep features extracted from a general deep CNN model. Large intra-class variation usually leads to erroneous identification on a pair of faces from the same individual at different ages. The mixed features (age features and identity features) potentially reduce the robustness of recognizing cross-age faces. To address this, we propose a new approach called orthogonal embedding CNNs. Below we first walk through the problem of deep AIFR in detail. 
%We start by depicting the problem of deep AIFR as:

Given an observed Fully-Connected (FC) feature $x$ extracted from the deep CNN model, we decompose it into two components (vectors). One is identity-related component $x_{id}$ and the other is age-related component $x_{age}$. Thus, after removing $x_{age}$ from $x$, we can obtain $x_{id}$ that is supposed to be age-invariant. 
Recent works \cite{hfa,LFCNN,AECNN} use a linear combination to model $x_{age}$ and $x_{id}$ as the solution. In this paper, we propose a new approach to model $x_{age}$ and $x_{id}$ in an orthogonal manner with deep convolutional neural networks. Inspired by A-Softmax \cite{sphereface}, where features of different identities are discriminated by different angles, we decompose feature $x$ in spherical coordinate system $x_{sphere} = \{r; \phi_1, \phi_2, ..., \phi_{n} \}$. The angular components $\{\phi_1, \phi_2, ..., \phi_{n}\}$ represent identity-related information, and the rest radial component $r$ is used to encode age-related information. Formally, $x \in R^{n}$ is decomposed under $x_{sphere}$ as
\begin{equation}\label{2}
x = x_{age} \cdot x_{id},
\end{equation}
where $x_{age}=||x||_2$, and $x_{id}=\{\frac{x_1}{||x||_2}, \frac{x_2}{||x||_2}, ..., \frac{x_n}{||x||_2}\}$, with $||x_{id}||_2=1$. Here $||.||_2$ represents for $L_2$ norm, and $x_n$ is the n-th component of $x$. For convenience, we will use $n_x$ to represent for $||x||_2$ and $\tilde x$ for $\frac{x}{||x||_2}$.
%%Though the decomposition is motivated by spherically distributed face features as pointed out by \cite{sphereface}, we found that the identity feature $x_{id}$ fits perfectly into the face matching process where cosine distance is used to calculate the similarity score between a pair of testing face features. With our decomposition, the identity component $x_{id}$ naturally has unit $L_2$ norm, which avoids the necessity of feature normalization at the testing stage.

\subsection{Multi-Task Learning}

\begin{figure*}[t]
\begin{center}
   \includegraphics[width=1 \linewidth, keepaspectratio]{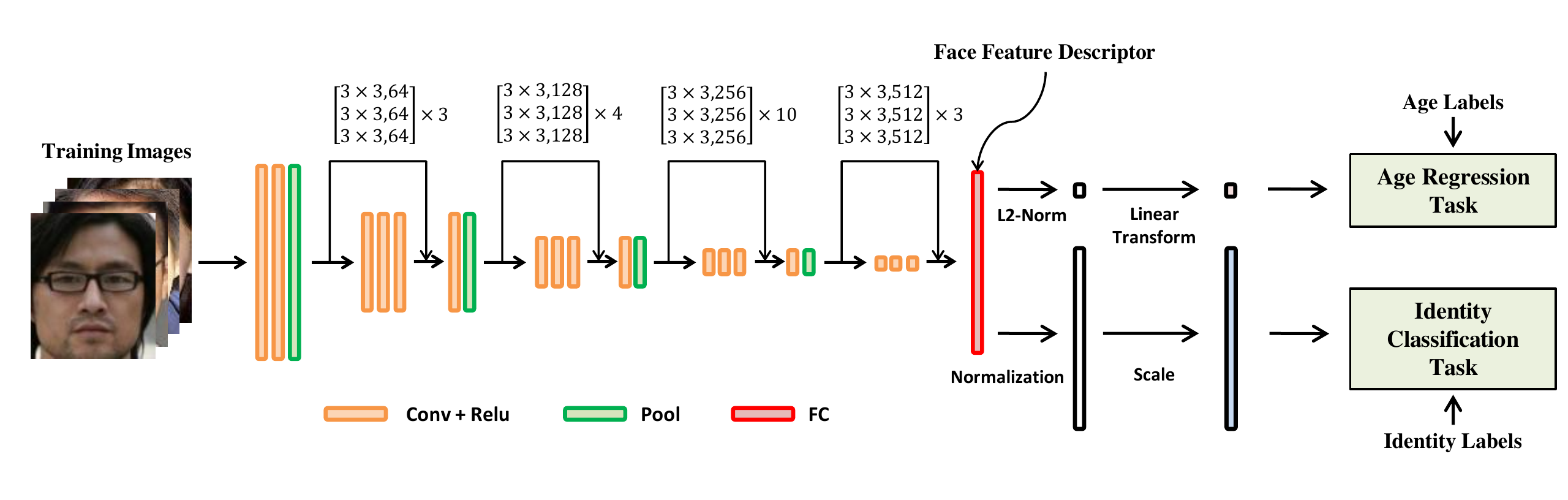}
\end{center}
   \caption{The proposed ResNet-Like CNN architecture.}
\label{fig::2}
\end{figure*}

According to Equation \ref{2}, feature $x$ output from the last FC layer is decomposed into $x_{age}$ and $x_{id}$. In this part, we describe a multi-task based learning algorithm to jointly learn these features. An overview of the proposed CNN model is illustrated in Figure \ref{fig::2}.

\textbf{Learning age-related component.}
In order to dig out the intrinsic clues of age information, we utilize an age estimation task to learn the relationship between the component $x_{age}$ ($n_x$) and the ground truth of age. For simplicity, linear regression is adopted to the age estimation task, and the regression loss can be formulated as follows:
\begin{equation}\label{3}
{L_{age}} = \frac{1}{{2M}}\sum\limits_{i = 1}^M {||f({n_{{x_i}}}) - {z_i}||_2^2}
%{L_{age}} = \frac{1}{{2M}}\sum\limits_{i = 1}^M h(i){||f({n_{{x_i}}}) - {z_i}||_2^2}
\end{equation}
where $n_{x_i}$ is the $L_2$ norm of the i-th embedding feature $x_i$, $z_i$ is the corresponding i-th age label.%, and $f(x) = k \cdot x + b$ is a simple one-dimensional polynomial function to apply transformation on $n_{x_i}$.
$f(x)$ is a mapping function aimed to associate $n_{x_i}$ and $z_i$. Since the $L_2$ norm $n_{x_i}$ is a scalar, we use 
linear polynomial $f(x) = k \cdot x + b$ as the mapping function. We also explored other more complicated functions such as non-linear multi-layer perceptron network, but they did not perform as well as a simple linear transformation. We believe this is because a more complicated model overfits the underlying feature which is one-dimensional here.

\textbf{Learning identity-related component.}
When performing face verification or identification, $\tilde x$ is the only  part which participates in the final similarity measure. Thus, the identity-related component $x_{id}$ should be as discriminative as possible. Following the recent state-of-the-art GFR algorithm A-Softmax \cite{sphereface}, we use a similar loss function to increase classification margin between different training persons in angular space:
\begin{equation}\label{4}
{L_{id}} = \frac{1}{{M}}\sum\limits_{i = 1}^M { - \log (\frac{{{e^{s \cdot \psi ({\theta _{{y_i},i}})}}}}{{{e^{s \cdot \psi ({\theta _{{y_i},i}})}} + \sum\nolimits_{j \ne {y_i}} {{e^{s \cdot \cos ({\theta _{j,i}})}}} }})}
\end{equation}
in which $\psi (.)$ is defined as $\psi ({\theta _{{y_i},i}}) = {( - 1)^k}\cos (m{\theta _{{y_i},i}}) - 2k$,
${\theta _{{y_i},i}}$ is the angle between the i-th feature $\tilde x_i$ and label $y_i$'s weight vector, ${\theta _{{y_i},i}} \in [\frac{{k\pi }}{m},\frac{{(k + 1)\pi }}{m}]$, and $k \in [0, m - 1]$. $m \ge 1$ is an integer hyper-parameter that controls the size of angular margin, and $s>0$ is an adjustable scale factor introduced to compensate the learning of Softmax. From the geometric perspective, Equation \ref{4} adds a constraint which guarantees the angle of the feature $x$ with its corresponding weight vector should less than $\frac{1}{m}$ of the angle between the feature $x$ and any other weight vectors. Consequently, the margin between two arbitrary classes can be increased. Compared with the original A-Softmax, Equation \ref{4} replaces $L_2$ norm of $\tilde x$ with an adjustable scalar factor $s$. In our model,  according to Equation \ref{2}, $||\tilde x||_2$ is always equal to 1. Thus, it is necessary to introduce an extra free variable to compensate for the loss of $L_2$ norm.

Overall, the two losses are combined to a multi-task loss for jointly optimizing, as below:
\begin{equation}\label{5}
\begin{array}{c}
L = {L_{id}} + \lambda {L_{age}}
\end{array}
\end{equation}
where $\lambda$ is a scalar hyper-parameter to balance the two losses.
Equation \ref{5} is used to guide the learning of our CNN model in the training phase. 
%In the training phase, we empirically set the hyper-parameters $m$, $s$, $\lambda$ to 4, 32, 0.01, respectively. Standard stochastic gradient descent (SGD) is applied for parameter updating. 
In the testing phase, only the identity-related component $x_{id}$ is used for the AIFR task.

\begin{figure*}[t]
\begin{center}
   \includegraphics[width=1 \linewidth, keepaspectratio]{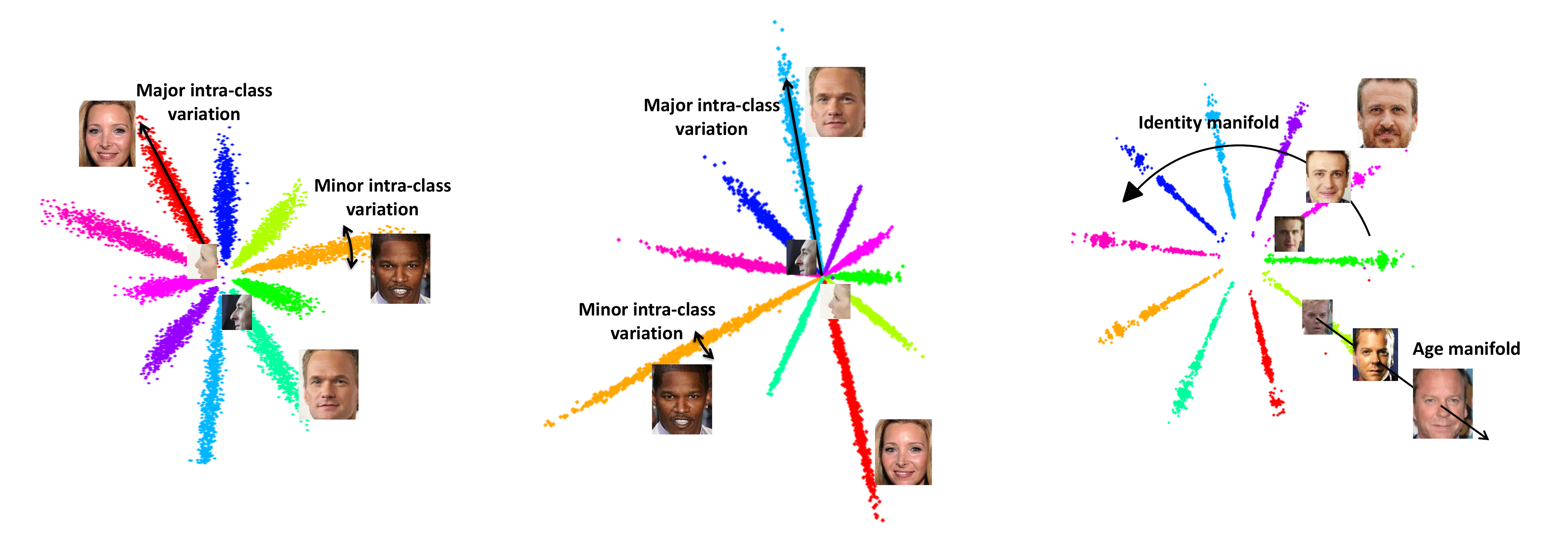}
\end{center}
   \caption{Visualization of deep features learned with Softmax (Left), A-Softmax (Middle) and the proposed algorithm (Right). It is noteworthy that only 10 individuals are used to train CNN models, and the output dimension is set to 2. Colors are used to distinguish identities, and placement of face images is based on the corresponding features.}
\label{fig::3}
\end{figure*}

\subsection{Discussion}
\textbf{Compared with HFA based AIFR methods.} The HFA based AIFR methods \cite{hfa,mefa,LFCNN} suggest modeling the identity-related component and age-related component of features by the simple linear combination. Specifically, given a feature $x$, the HFA based methods decompose the $x$ as $x = m + U{x_{age}} + V{x_{id}} + \varepsilon$,
where $m$ is the mean feature regarding identity-related component, $\varepsilon$ is the additional noise and $U, V$ are the transformation matrices for identity-related component $x_{id}$ and age-related component $x_{age}$ respectively. The major advancements of the proposed approach over the HFA based methods are described in the following aspects:
Firstly, the proposed approach revises the decomposition of $x$ in the HFA based methods to the multiplication of hidden components $x_{id}$ and $x_{age}$, which is more intuitive and concise to model the unrelated components with less extra hyper-parameters.
Secondly, we explicitly project the identity features on a hypersphere to match the cosine similarity measurement for effectively combining the improvement strategies based on the Softmax loss and the margin of decision boundaries.
Thirdly, the HFA based methods have to iteratively run the EM algorithm in contrast to our approach which jointly trains the network in the desirable end-to-end manner of feature learning.
For the foregoing reasons, our method is more recommendable to be embedded into CNN framework for the purpose of learning age-invariant features, as supported by our experimental results.

\textbf{Compared with SphereFace.}
SphereFace \cite{sphereface} introduces A-Softmax loss to learn the angular margin between identities for GFR. Though we train the identity-related component with a loss function similar to A-Softmax, the proposed algorithm takes advantage of the age information to explicitly train age-related component with an additional age regression task (Equation \ref{3}). To intuitively investigate the impact by introducing such additional age regression task, we construct a toy example to compare features learned by Softmax, A-Softmax and our proposed algorithm. Specifically, we train CNN models with 10 individuals and set the output dimension of feature $x$ as 2. For simplicity we let $f(x) = x$ (see Equation \ref{3}) in this case. Figure \ref{fig::3} is the visualization for training features. Based on this example, we conclude that: (1) features of different persons are discriminated mostly by angles, which intuitively justifies our decomposition design; (2) both A-Softmax and the proposed algorithm have noticeably larger classification margins than Softmax, as a result of the A-Softmax loss;  (3) most importantly, for our model age of a person is reflected in radial direction (e.g. larger $L_2$ norms for older faces), while the other two models do not have this property. We believe this property further constrains the training problem, which reduces the risk of over-fitting and consequently leads to superior performance for AIFR.

\textbf{Generalization of Our Approach.}
One of the noticeable highlights of the proposed algorithm is its generalization capability.
Intuitively, our method is specifically designed to fit cross-age training data.
However, the experimental results surprisingly unfold the excellent performance of the proposed method even trained with general training data (as shown in Section \ref{sec4_4}).
Furthermore, as the objective of the algorithm is to generate identity-related features, the proposed algorithm is not only suitable for AIFR but also for GFR.
Finally, The age component can be easily generalized to any other common component such as pose, illumination, emotion, etc.

\section{Large-Scale Cross-Age Face Dataset (CAF)}
\label{sec4}

\begin{figure*}[t]
  \centering
  % Requires \usepackage{graphicx}
  %An example case which 类间距离 小于 类内距离
  \subfigure[]
  {
    \label{dataset_examples}
    \includegraphics[width=0.4\linewidth, keepaspectratio]{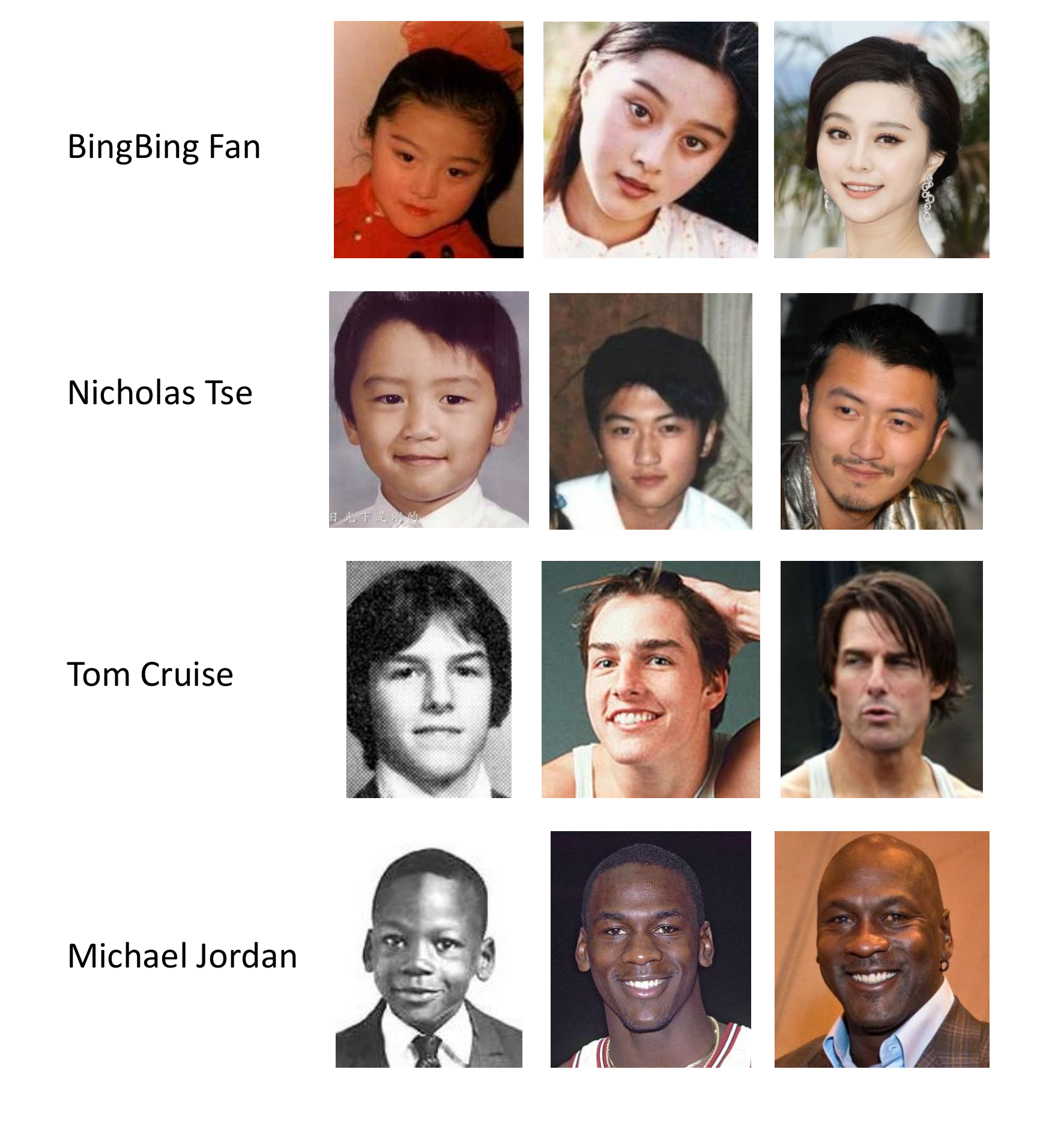}
  }
  \subfigure[]
  {
    \label{dataset_dist}
    \includegraphics[width=0.5 \linewidth, keepaspectratio]{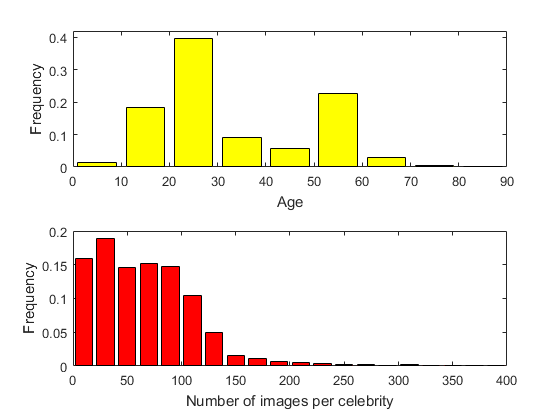}
  }
  %The major challenge of AIFR. The large intra-class variations 使得识别人脸的难度大大提高。
  \caption{Overview of the CAF dataset. (a) Example images of CAF. Note that since our images are collected from Internet, CAF not only varies in ages but also in poses, races, etc. (b) The distribution of CAF. Top: The distribution of the number of different ages. Bottom: The distribution of the number of different identities. }
  \label{fig:1}
\end{figure*}

In order to further motivate the development of AIFR and enrich the capability of the current model,
a dataset with a large age gap is urgently needed. Besides, the dataset size should be large enough to avoid overfitting. To this end, we collect a new dataset with a large number of cross-age celebrities' faces, named large-scale Cross-Age Face dataset (CAF).

\subsection{Dataset Collection}
To build the cross-age celebrity dataset, it is inevitable to collect celebrity's name to form a list. The collected names in the list come from multiple sources such as IMDB, Forbes celebrity list, child actors name list from Wikipedia, etc. This guarantees the comparatively large age gap in the later data collection. Next, we iteratively search the name in the list by the Google Search Engine. Each searching term has been thresholded to a certain number, that is, we keep the name in the list if the number of responses exceeds a certain threshold, which ensures the sufficient number of data for each celebrity. Moreover, to the best of our knowledge, the current public cross-age datasets have very limited Asian individuals. For the purpose of increasing the diversity of our cross-age dataset, we collect a large number of Asian celebrities. After filtering the name list, we download the face images on several commercial image search engine (such as Google, Baidu) querying by the celebrity's name companied with several keywords like \emph{yearbook}, \emph{past and now}, \emph{childhood}, \emph{young}, \emph{from young to old}, etc, to obtain the face images with different age categories. The data cleaning is performed thereafter. Specifically, we apply face detection algorithm MTCNN \cite{mtcnn} to filter the images without any faces, then manually wipe off the near-duplicates and false face images (faces do not belong to that celebrity). Finally, we delete some of the images that have a large proportion in a certain age category to keep the age distribution more balanced.

%TODO：考虑替换AgeDB和其他dataset的顺序，imagenum从大到小排列
\begin{table}[t]
%\footnotesize
\begin{center}
\begin{tabular}{|c|c|c|c|c|c|c|}
\hline
Dataset 	& CAF 	& IMDB-WIKI	\cite{imdb2} & CACD \cite{cacd} 		& MORPH \cite{morph}   & AgeDB \cite{agedb} & FG-NET \cite{fgnet}			\\
\hline\hline
\# Images 	& 313K 		& 523K	 	& 163K 		& 78K &	16K		& 1K 				\\
\# Subjects & 4,668		& 20,284 	& 2,000		& 20,000&	568	& 82				\\
Noise-free	& Yes		& No     		& Yes		& Yes	&	Yes	& Yes				\\
\hline
\end{tabular}
\end{center}
\caption{Comparison over cross-age datasets.}
\label{dataset_comp}
\end{table}

\subsection{Dataset Statistics}
Following the above labeling and cleaning process, we construct a cross-age face dataset which totally includes about 313,986 face images from 4,668 identities. Each identity has approximately 80 face images.
All of these images have been carefully and manually annotated.
Example images of the dataset are shown in Figure \ref{dataset_examples}.
Considering the lack of exact age information, we utilize the public pre-trained age estimation model DEX \cite{imdb2} to predict the rough age label for each face image. Figure \ref{dataset_dist} shows the distribution histogram of CAF. One can observe our data are well-distributed in every possible age category. Table \ref{dataset_comp} fairly compares our dataset with existing released cross-age datasets. It is clear that except IMDB-WIKI \cite{imdb2}, we have the comparatively largest scale in terms of the number of pictures and the number of individuals. Furthermore, as IMDB-WIKI is collected by automatically online crawling, some of the downloaded data might be redundant and noise-severe.
Superior to IMDB-WIKI, CAF has minimized the noise data by manually annotating.

\section{Experiments}
\label{sec5}
For a direct and fair comparison to the existing work in this field, we evaluate our approach on existing public-domain cross-age face benchmark datasets MORPH Album 2\cite{morph}, CACD-VS \cite{cacd} and FG-NET \cite{fgnet}. We also evaluate our algorithm on LFW \cite{lfw} for verifying the generalization performance on GFR.

\subsection{Implementation Details}
The training set is composed of two parts: a cross-age face dataset and a general face dataset (without cross-age face data). The cross-age face dataset that we use is the collected CAF dataset introduced in Section \ref{sec4} while the general face dataset consists of three public face datasets: CASIA-WebFace \cite{webface}, VGG Face \cite{vggface} and celebrity+ \cite{celeb}.
The same identities appeared in different datasets are carefully merged together. Since our testing dataset contains MORPH, CACD-VS, FG-NET, and LFW, we have excluded these data from the training set.
Finally, our training set contains 1,765,828 images with 19,976 identities in total, which includes 313,986 cross-age face images with 4,668 identities and 1,451,842 general face images with 17609 identities respectively.
In addition, the age label predicted from the public pre-trained age estimation model DEX \cite{imdb2} is treated as the regression target of Euclidean loss.
Prior to training stage, we perform the same pre-processing on both training set and testing set: Using MTCNN \cite{mtcnn} to detect the face and facial key points in images, then applying similarity transformation to crop the face patch to 112$\times$96 pixels according to the 5 facial key points (two eyes, nose and two
mouth corners), finally normalizing the cropped face patch by subtracting 127.5 then divided by 128. The proposed loss in Equation \ref{4} serves as the supervisory signal of identity classification. In terms of the age branch, we use Euclidean loss function to guide the network to learn the age label. 
%Unless specified, the hyper-parameters $m$, $s$, $\lambda$ mentioned in Equation \ref{4} and \ref{5} are set to 4, 32, 0.01, respectively.
The hyper-parameters $m$, $s$ mentioned in Equation \ref{4} and \ref{5} are set to 4, 32 according to the recommendations of \cite{sphereface,normface}. For the factor $\lambda$, we empirically selected an optimal value 0.01 to balance the two losses.
%There are primarily three hyper-parameters introduced in this work: margin $m$, scale factor $s$ mentioned in equation \ref{4} and \ref{5}, and loss balancing factor $\lambda$. We set $m=4$ and $s=32$ according to the recommendations of \cite{sphereface,normface}. For $\lambda$, we empirically tested a few values (0.001, 0.01 and 0.1) on development data and selected an optimal one (0.01).
All models are trained with Caffe \cite{caffe} framework and optimized with stochastic gradient descent (SGD) algorithm. Training batch size is set to 512 and the number of iterations is set to 21 epochs. The initial learning rate is set to 0.05 and the training process adaptively decreases the learning rate 3 times when the loss becomes stable (roughly at the 9-th, 15-th and 18-th epoch).

\subsection{Experiments on the MORPH Album 2 Dataset}
Following \cite{hfa,mefa,lps_hfa,LFCNN}, in this study we use an extended version of MORPH Album 2 dataset \cite{morph} for performance evaluation. It has 78,000 face images of 20,000 identities in total. The data has been split into training and testing set. The training set contains 10,000 identities.
%For quantitatively measuring the CAF dataset on performance-wise, we neglect the provided MORPH Album 2 training set.
The rest of 10,000 identities belong to testing set where each identity has 2 photos with a large age gap. The testing data have been divided into gallery set and probe set. We follow the testing procedure given by \cite{hfa} to evaluate the performance of our algorithm. We set up several schemes for comparison including: (1) \textbf{Softmax}: the CNN-baseline model trained by the original Softmax loss, (2) \textbf{A-Softmax}: the CNN-baseline model guided by the A-Softmax loss, (3) \textbf{OE-CNNs}: the proposed approach, and (4) other recently proposed top-performing AIFR algorithm in the literatures.

Firstly, we compare the proposed approach to baseline algorithms that are most related to the proposed algorithm to demonstrate its effectiveness. Table \ref{table:dataset_exp} compares the rank 1 identification rates testing on 10,000 subjects of Morph Album 2 over Softmax, A-Softmax, and OE-CNNs, with and without CAF dataset. 
As shown in the table, The proposed \textbf{OE-CNNs} significantly outperforms both Softmax and A-Softmax under both settings. Specifically, though we've used similar loss function with A-Softmax for training the identity-related features, \textbf{OE-CNNs} noticeably improves the performance of A-Softmax, which confirms the effectiveness of our features decomposition method for AIFR. Note that, all compared networks have the same base network (from input to FC layer). When comparing performances trained with and without CAF dataset, we can see that with CAF the identification rate improves consistently for all systems, which confirms that the CAF dataset is valuable to AIFR research.

Secondly, for ensuring a fair comparison with other methods, we neglect the CAF dataset and conduct an experiment with the same training data as related work \cite{LFCNN} has used. Specifically, WebFace \cite{webface}, celebrity+ \cite{celeb} and CACD \cite{cacd} form the training set to train a CNN base model. The trained model is later fine-tuned with Morph training data.
Table \ref{tabel:MORPH-EXP} depicts our result compared with other methods.
There are conventionally two evaluation schemes on Morph benchmark: testing on 10,000 subjects or 3,000 subjects. For fairly comparing against other methods, we evaluate the proposed OE-CNN approach on both schemes.
As can be seen in Table \ref{tabel:MORPH-EXP}, the OE-CNN approach shows its capability by substantially outperforming all other methods in both two evaluation schemes. Particularly, our method surpasses the LF-CNN model by 1.0\% and AE-CNN model by 0.5\%, which is an outstanding improvement on the accuracy level above 98\%. 

\begin{table}[t]
%\small
\begin{center}
\begin{tabular}{|c|c|c|}
\hline
Training Dataset							& Method 	& \tabincell{c}{Rank-1\\Identification Rates} \\
\hline\hline
Public datasets 				& Softmax  		& 94.84\% \\
Public datasets					& A-Softmax   	& 96.27\% \\
Public datasets					& \textbf{OE-CNNs} 		& \textbf{97.46\%} \\
\hline\hline
Public datasets + CAF 			& Softmax  		& 95.49\% \\
Public datasets + CAF			& A-Softmax   	& 96.59\% \\
Public datasets + CAF			& \textbf{OE-CNNs}   		& \textbf{98.57\%} \\
\hline
\end{tabular}
\end{center}
\caption{Performance comparisons of different baselines on Morph Album 2.}
\label{table:dataset_exp}
\end{table}

\begin{table}[t]
%\small
\begin{center}
\begin{tabular}{|c|c|c|}
\hline
Method	& \tabincell{c}{\#Test Subjects}			& \tabincell{c}{Rank-1\\Identification Rates}\\
\hline\hline
HFA \cite{hfa} 		& 10,000			& 91.14\% \\
CARC \cite{cacd}	& 10,000				& 92.80\% \\
MEFA \cite{mefa}	& 10,000				& 93.80\% \\
MEFA+SIFT+MLBP \cite{mefa}	& 10,000	& 94.59\% \\
LPS+HFA \cite{lps_hfa}	& 10,000		& 94.87\% \\
LF-CNNs \cite{LFCNN}	& 10,000			& 97.51\% \\
\textbf{OE-CNNs}		&	 10,000	& \textbf{98.55\%} \\
\hline
GSM \cite{gsm}		& 3,000			& 94.40\% \\
AE-CNNs \cite{AECNN}	& 3,000			& 98.13\% \\
\textbf{OE-CNNs} 				& 3,000			& \textbf{98.67\%} \\
\hline
\end{tabular}
\end{center}
\caption{Performance comparisons of different approaches on Morph Album 2.}
\label{tabel:MORPH-EXP}
\end{table}

% \begin{table}
% \begin{center}
% {
% \footnotesize
% \begin{tabular}{c|c|c|c|c|c}
% \hline
% Method 	& \cite{geng2013facial} (2013) & \cite{lu2015cost} (2015) & \cite{liu2017group} (2017) & OE-CNNs & OE-CNNs*\\
% \hline\hline
% MAE & 5.67 & 4.37 & 3.25 & 5.28 & 3.66\\
% \hline
% \end{tabular}
% }
% \end{center}
% \caption{MAE Comparisons of different approaches on Morph Album 2.}
% \label{table:mae}
% \end{table}

% \textcolor{red}{Moreover, we evaluate the age regression task with mean absolutely errors (MAE) on Morph Album 2 against systems specifically designed for age estimation. Table \ref{table:mae} indicates that the most recent state-of-the-art methods performs better than ours. This is because our method is specifically optimized for face recognition rather than age estimation, where we only use 1-d radius to model age information. Besides, the age labels of the training data are generated from the output of the DEX model \cite{imdb2}, which may be biased. Note that if we treat the predicted results of DEX as ground-truths for testing images, the MAE reduced significantly (mentioned as OE-CNNs*). We believe that OE-CNNs certainly manage to achieve better age estimation performance with high-quality and accurate-labeling data.   
%}

\subsection{Experiments on the CACD-VS Dataset}
CACD dataset comprises comprehensively 163,446 images from 2,000 distinct celebrities. The age ranges from 10 to 62 years old. This dataset collects the celebrity's images with the effect of various illumination condition, different poses and makeup, which can effectively reflect the robustness of the AIFR algorithm. CACD-VS is a subset of CACD which is picked from CACD to composes 2,000 pairs of positive sample and 2,000 pairs of negative samples, and 4,000 pairs of samples in total. We follow the pipeline of \cite{cacd} to calculate the similarity score of all sample pairs and the ROC curves and its corresponding AUC. We take 9 folds from 10 folds that have already been separated officially to compute threshold references and use this threshold to evaluate on the rest of 1 fold. By repeating this procedure 10 times, we finally calculate the average accuracy as another measure.

\begin{figure}[t]
\begin{center}
   \includegraphics[width=0.7 \linewidth, keepaspectratio]{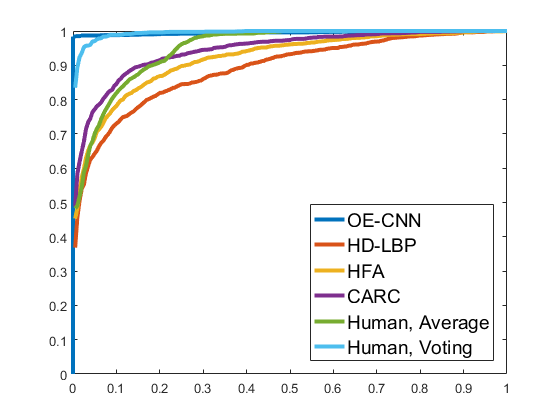}
\end{center}
   \caption{ROC comparisons of different approaches on CACD-VS.}
\label{fig:cacd-roc}
\end{figure}

\begin{table}[t]
\begin{center}
\begin{tabular}{|c|c|c|}
\hline
Method 													& Acc.					& AUC.\\
\hline\hline
High-Dimensional LBP \cite{chen2013blessing}		& 81.6\%				& 88.8\% \\
HFA \cite{hfa} 									& 84.4\%				& 91.7\% \\
CARC \cite{cacd}									& 87.6\%				& 94.2\% \\
LF-CNNs \cite{LFCNN}								& 98.5\%				& 99.3\% \\
Human, Average \cite{cacd2}						& 85.7\%				& 94.6\% \\
Human, Voting \cite{cacd2} 						& 94.2\%				& 99.0\% \\
\hline\hline
Softmax										& 98.4\% 				& 99.4\% \\
A-Softmax										& 98.7\% 				& 99.5\% \\
\textbf{OE-CNNs}											& \textbf{99.2\%} 		& \textbf{99.5\%} \\
\hline
\end{tabular}
\end{center}
\caption{Performance comparisons of different approaches on CACD-VS.}
\label{tabel:CACD-EXP}
\end{table}

The results of all the baselines are shown in Table \ref{tabel:CACD-EXP} and Figure \ref{fig:cacd-roc}.
As illustrated, the proposed OE-CNN approach significantly outperforms
all the other baselines. Furthermore, our approach also surpasses the human-level performance, which demonstrates the effectiveness of our proposed age-invariant deep features.

\subsection{Experiments on the FG-NET Dataset}
\label{sec4_4}

The FG-NET dataset consists of 1,002 pictures from 82 different identities, each identity has multiple face images with huge variability in the age covering from child to elder.
%There are plenty of cross-age images for each identity.
Following the evaluation protocols of Megaface challenge 1 (MF1) \cite{mf1} and Megaface challenge 2 (MF2) \cite{mf2} we employ the 1 million images from Flickr as the distractor set.
Particularly, under the small protocol of MF1, we reduce our training data to 0.5 million images from 12,073 identities in the training phase.
The cross-age face images in FG-NET servers as the probe set in which a probe image is compared against each image from distractor set. We evaluate the rank-1 performance of the presented algorithm under the protocols of MF1 and MF2, as shown in Table \ref{tabel:FGNET-EXP} and Table \ref{tabel:FGNET-EXP2}, respectively.
%Under the protocol of MF1, our algorithm beats most of other published methods except FaceNet \cite{facenet}. Nevertheless, FaceNet is trained on 200 million face data in contrast to 1.76 million of our model.

Under the small protocol of MF1, the proposed method not only obtains a significant performance improvement over Softmax and A-Softmax baseline but also surpasses the existing methods (including a specific age-invariant method TNVP \cite{TNVP}) by a clear margin.
Under the protocol of MF2, all the algorithms need to be trained using the same training dataset (which does not involve the cross-age training data) provided by MF2 organizer. It is encouraging to see that our algorithm also outperforms all other methods with a large margin, which strongly proves the effectiveness of our algorithm on AIFR.

\begin{table}[t]
\small
\begin{center}
\begin{tabular}{|c|c|c|}
\hline
Method 			& Protocol 											& \tabincell{c}{Rank-1\\Identification Rates}\\
\hline\hline
FUDAN-CS\underline{ }SDS \cite{fudan}	& Small	& 25.56\% \\
SphereFace \cite{sphereface}		& Small							& 47.55\% \\
TNVP \cite{TNVP}			& Small						& 47.72\% \\
%SIAT\underline{ }MMLAB		& Small						& 55.30\% \\
%FaceNet v8 \cite{facenet} & Large						& 74.59\% \\
%\hline\hline
%Softmax					& Large						& 44.09\% \\
%A-Softmax				& Large							& 50.69\% \\
\hline\hline
Softmax					& Small						& 35.11\% \\
A-Softmax				& Small						& 46.77\% \\
%单patch
OE-CNNs (single-patch)		& Small						& 52.67\% \\
%多patch
\textbf{OE-CNNs (3-patch ensemble)}			& Small						& \textbf{58.21\%} \\
% \textbf{OE-CNNs}			& Large						& \textbf{59.08\%} \\
\hline
\end{tabular}
\end{center}
%\caption{Performance of different published approaches on FG-NET. Note that the experiments are conducted under the protocol of MF1 \cite{mf1}.}
\caption{Performance comparisons of different approaches under the protocols of MF1 \cite{mf1} on FG-NET.}
\label{tabel:FGNET-EXP}
\end{table}

\begin{table}[t]
\small
\begin{center}
\begin{tabular}{|c|c|c|}
\hline
Method 			& Protocol 											& \tabincell{c}{Rank-1\\Identification Rates}\\
\hline\hline
GRCCV	& Large	& 21.04\% \\
NEC	& Large	& 29.29\% \\
3DiVi	& Large	& 35.79\% \\
GT-CMU-SYSU	& Large	& 38.21\% \\
%\hline\hline
%Softmax					& Large						& 44.09\% \\
%A-Softmax				& Large							& 50.69\% \\
\hline\hline
\textbf{OE-CNNs (single-patch)}			& Large						& \textbf{53.26\%} \\
\hline
\end{tabular}
\end{center}
\caption{Performance comparisons of different approaches under the protocol of MF2 \cite{mf2} on FG-NET.}
\label{tabel:FGNET-EXP2}
\end{table}

\subsection{Experiments on the LFW Dataset}
LFW is a very famous benchmark for general face recognition. The dataset has 13,233 face images from 5,749 subjects acquiring from the arbitrary environment. We experiment our algorithm on LFW following the official unrestricted with labeled outside data protocol. We test our model on 6,000 face pairs. The training data are disjoint from the testing data. Table \ref{tabel:LFW-EXP} exhibits our results. One can see that the proposed OE-CNN approach achieves comparable performance without any ensemble trick to the state-of-the-art approaches, which demonstrates the excellent generalization ability of the proposed approach. Additionally, after we expand the training dataset to 1.7M (including CAF dataset), the performance of OE-CNNs further improves to 99.47\%, which also proves that our CAF dataset is not only valuable for AIFR but also helpful for GFR.  
%将数据增加至1.7M之后（结合CAF数据），OE-CNNs的性能提升至99.47%，这进一步说明了我们的CAF数据集不仅在跨年龄人脸识别上的价值，同时也对通用人脸识别有所帮助。
%To evaluate the generalization performance of LF-CNNs, we further conduct an experiment on the famous LFW dataset. This dataset contains 13,233 face images from 5749 different subjects, collecting from uncontrolled conditions.Following the unrestricted with labeled outside data protocol, we train on the outside dataset and test on 6,000 face pairs. People overlapping between the outside training data and the LFW testing data are excluded. We respectively train 25 networks with 25 different image patches,and concatenate the output features from these networks into a long feature vector. We then apply PCA on the longfeature vector to obtain a compact feature vector for classifi-cation. In Table 5 we compare our results against the recent

\begin{table}[t]
\footnotesize
\begin{center}
\begin{tabular}{|c|c|c|c|c|}
\hline
\multicolumn{2}{|c|}{Method}  & Images			& Networks			& Acc.	\\
\hline\hline
\multirow{5}{*}{\tabincell{c}{General\\Approaches}}
& DeepFace \cite{deepface} & 4M  & 3 & 97.35\%\\
& FaceNet \cite{facenet} & 200M  & 1 & 99.65\%	\\
& DeepID2+ \cite{deepid2plus}  & -  & 25 & 99.47\%	\\
& Center Loss \cite{centerloss}  & 0.7M  & 1 & 99.28\%	\\
& SphereFace \cite{sphereface}  & 0.5M  & 1 & 99.42\%	\\
\hline\hline
\multirow{3}{*}{\tabincell{c}{Cross-Age\\Approaches}}
& LF-CNNs \cite{LFCNN} & 0.7M  & 1 & 99.10\%   \\
%& LF-CNNs \cite{LFCNN} & 0.7M  & 25 & 99.50\%   \\
&\textbf{ OE-CNNs}  & 0.5M  & 1 & \textbf{99.35\%}	\\
&\textbf{ OE-CNNs}  & 1.7M  & 1 & \textbf{99.47\%}	\\
%\textbf{OE-CNNs}			& \textbf{99.1\%} 		& \textbf{99.5\%} \\
\hline
\end{tabular}
\end{center}
\caption{Performance comparisons of different approaches on LFW.}
\label{tabel:LFW-EXP}
\end{table}

\section{Conclusion}
AIFR is a remained challenging computer vision task on account of the aging process of the human. Inspired by pioneering work and the observation of hidden components, this paper proposes a novel approach which separates deep face feature into the orthogonal age-related component and identity-related component to improve AIFR.
The highly discriminative age-invariant features can be consequently extracted from a multi-task deep CNN model based on the proposed approach. Furthermore, we build a large cross-age celebrity dataset named CAF that is both noise-free and vast in the number of images. As a part of training data, CAF greatly boosts the performance of the models for AIFR.
Extensive evaluations of several face aging datasets have been done to show the effectiveness of our orthogonal embedding CNN (OE-CNN) approach.
More studies on how to incorporate the generative scheme and improve the discriminative scheme will be explored in our future work to benefit the AIFR community.

% In this paper, we propose a novel approach for AIFR which separates deep face feature into the orthogonal age-related component and identity-related component.
% As a result, high discriminative age-invariant features can be extracted from a multi-task supervised deep CNN model. Furthermore, we build a large cross-age celebrity dataset named CAF that is both noise-free and vast in number of images. CAF greatly boosts the performance of the proposed models as a part of training data.
% Extensive evaluations over several face aging benchmark datasets have been done to show the effectiveness of our orthogonal embedding CNN (OE-CNN).
% More researches on how to incorporate generative scheme into our approach will be explored in our future work to benefit the AIFR community.

\clearpage

\bibliographystyle{splncs04}
\bibliography{egbib}
\end{document}